\pdfoutput=1

\documentclass[11pt]{article}

\usepackage{emnlp2021}

\usepackage{times}
\usepackage{latexsym}
\usepackage{times}
\usepackage{latexsym}

\usepackage{algorithm}  
\usepackage{algorithmic}  
\usepackage{microtype}
\usepackage{amsfonts,amssymb}
\usepackage{graphicx}
\usepackage{float} 
\usepackage{subfigure}
\usepackage{microtype}
\usepackage{ulem}
\usepackage{color}
\usepackage{amsmath}
\usepackage{amssymb}
\usepackage[T1]{fontenc}

\usepackage[utf8]{inputenc}

\usepackage{microtype}

\title{BioCopy: A Plug-And-Play Span Copy \\ Mechanism in Seq2Seq Models}

\author{Yi Liu~\textsuperscript{1}, Guoan Zhang~\textsuperscript{2}, Puning Yu~\textsuperscript{3}, Jianlin Su~\textsuperscript{4}, Shengfeng Pan~\textsuperscript{4} \\
\textsuperscript{1}~RMIT University, Melbourne, Australia \\
\textsuperscript{2}~King's College London, the United Kingdom \\ 
\textsuperscript{3}~South China University of Technology, China \\
\textsuperscript{4}~Shenzhen Zhuiyi Technology Co., Ltd., China \\
yiliumelai@gmail.com, guoan.zhang@kcl.ac.uk, \\ ypn20010116@163.com, 
\{bojonesu,nickpan\}@wezhuiyi.com
}

\date{}

\begin{document}
\maketitle
\begin{abstract}
Copy mechanisms explicitly obtain unchanged tokens from the source (input) sequence to generate the target (output) sequence under the neural seq2seq framework. However, most of the existing copy mechanisms only consider single word copying from the source sentences, which results in losing essential tokens while copying long spans. In this work, we propose a plug-and-play architecture, namely BioCopy, to alleviate the problem aforementioned. Specifically, in the training stage, we construct a BIO tag for each token and train the original model with BIO tags jointly. In the inference stage, the model will firstly predict the BIO tag at each time step, then conduct different mask strategies based on the predicted BIO label to diminish the scope of the probability distributions over the vocabulary list. Experimental results on two separate generative tasks show that they all outperform the baseline models by adding our BioCopy to the original model structure. 
\end{abstract}

\section{Introduction}
 
Recent neural seq2seq systems have been successful in various NLP tasks which utilize an encoder to convert a source sentence into a fixed vector, and a decoder to generate a target sentence by using the semantic information amongst the fixed vectors. 
However, seq2seq suffers from the out-of-vocabulary and rare word problem\citep{luong-etal-2015-addressing,gulcehre-etal-2016-pointing} that words in source sentence are not able to be obtained to generate the target sentence. Due to this problem,~\citep{he2016deep,DBLP:journals/corr/SrivastavaGS15}, propose the so-called `copy mechanism' where it locates certain words in the input sentence and put these words into the target sequence. In their work, every output word can be generated either by predicting from the vocabulary or copying from the source sequence. Unfortunately, most of them copy the words from the source sentence to the target sentence in a word-by-word manner. However, in many cases, the copied words are generated consecutively from a span in the source sequence.  

In this paper, we propose a novel and portable copy mechanism to solve the problem of low accuracy when copying spans in the seq2seq framework. To implement this new copy mechanism, we propose a BIO-tagged strategy that annotates the target sequences with BIO tags by matching the longest common subsequence (LCS) with the source sequence. Therefore, the BIO tags are able to exactly locate the start and end position of every single span in target sequences. In the inference stage, we design a span extractor to determine copying spans from the source sequence. Specifically, the BIO tag is firstly predicted at each time step to indicate the position of the copied span, then we decide a copy algorithm to guide the span extractor by generating the eligible n-gram set.  

Since our method does not change the structure of the model itself, this BioCopy can be seen as a plug-and-play component, which is simple and effective enough to transfer and apply to any generative seq2seq framework. The experimental results in generative relation extraction and abstractive summarization tasks indicate the effectiveness of our proposed copy mechanism.

\section{Approach}

The conventional seq2seq framework generates one token at each time step. Therefore, given the predicted token sequence, the mode decoder predicts the current token by computing the probability distribution over the entire vocabulary list:
\begin{equation}
    y_t = p(y_t|y_{<t},x) 
\end{equation}
In our case, we add an extra sequence prediction task to the model decoder. Specifically, the former decoder just models the distribution on each token, but the current decoder predicts an additional label distribution:
\begin{equation}
\begin{split}
    y_t, z_t &= p(y_t,z_t|y_{<t},x) \\
             &= p(y_t|y_{<t},x)p(z_t|y_{<t},x)
\end{split}    
\end{equation}
where $z \in \{\text{B},\text{I},\text{O}\}$. The meaning is as follows:
\begin{itemize}

\item \textbf{B} indicates the current token is copied from the source sentence.
\item \textbf{I} indicates the current token is copied and forms a continuous fragment with the previous token from source sentence.
\item  \textbf{O} indicates the current token is not copied, but generated from the vocabulary list. 

\end{itemize}

\paragraph{BIO-tag Building} As we mentioned above, we utilize the supervised method to train the BIO tag $z$. In order to acquire the labels, we compute the longest common subsequence~\citep{10.5555/645723.666723} between the source sentence and target sentence. The token will be considered as copied from source sentence as long as it appears in the longest common subsequence, and different tags are assigned according to the specific meaning of BIO. In summary, in the training phase, besides the original token sequence prediction task, we also introduce one more sequence prediction task, of which the tags are all given. It is easy to implement and does not add any extra computational cost. The loss function $\mathcal{L}$ is defined as the following:
\begin{equation}
    \mathcal{L} = \frac{1}{\text{N}\text{T}} \sum^N \sum_i^T y_i^{'}\log y_i + z_i^{'}\log z_i
\end{equation}
where  $y^{'}$ and $z^{'}$ denotes gold label of token sequence and BIO-tag sequence, $T$ is the sequence length and $N$ is the batch size. $y$ and $z$ denotes predicted label.

\begin{figure*}
    \centering
    \includegraphics[width=\textwidth]{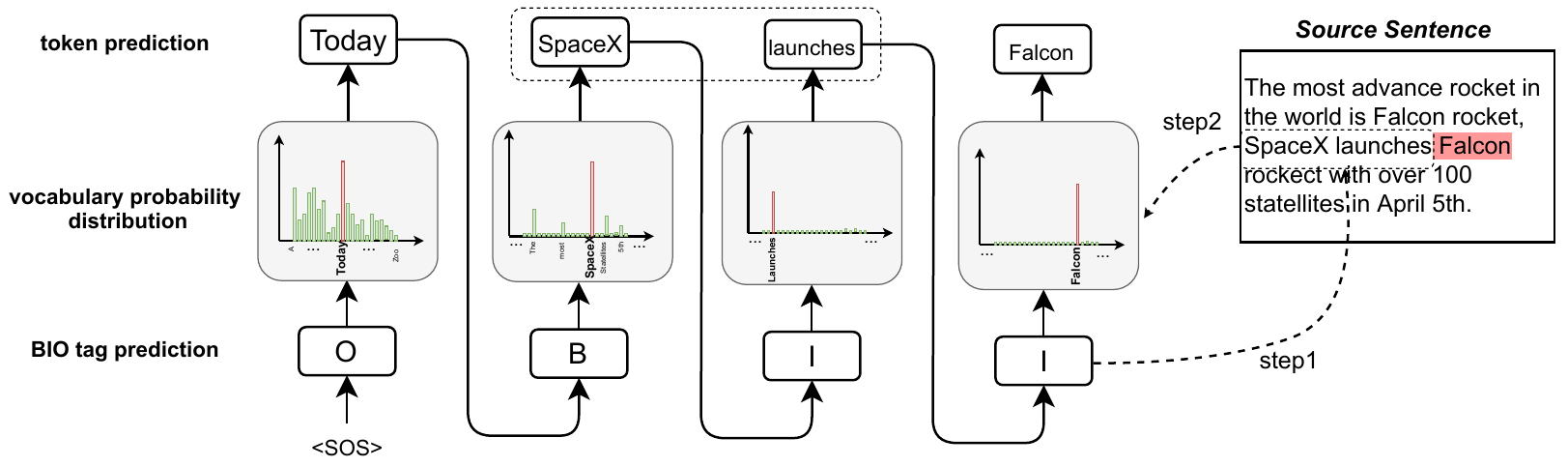}
    \caption{The workflow of model decoder in the inference stage: in order to predict the current token `Falcon', the model firstly predicts its BIO = I, which indicates the current token constitutes a 3-gram with the predicted tokens `SpaceX', `launches'. Then, we search the potential 3-grams from the source sentence, and the `Falcon' is the only valid candidate. Finally, the token probability distributions except `Falcon' are all masked as zero.}
    \label{fig:model_inference}
\end{figure*}

\paragraph{Inference}
In the inference stage, at each time step, we first predict the BIO-tag $z_t$, the result will be one of three circumstances: if $z_t = \text{O} $, we do not need to change anything. If $z_t = \text{B}$, we mask all the token probability distributions of which they are not in the source sentence. If $z_t = \text{I}$, it indicates the token sequence from current token to its nearest token with $z_t = \text{B}$ constitutes a consecutive n-gram from source sentence. Therefore, we mask all the tokens in the token probability distribution that cannot constitute the corresponding n-gram in the source sentence. In this way, the model decoder still generates tokens in a step-by-step manner, rather than generating a segment at one time step. According to utilise the mask operation, the tokens where their $z_t = \text{B}$ or $z_t = \text{O}$ are selected from a segment in the input sentence. A detailed example has been shown in Figure~\ref{fig:model_inference} for illustration.

It should be pointed out that the proposed copy mechanism can not only improve the model performance in terms of long-span extraction, but also ensures the consistency between the generated text and the original text, thereby avoiding professional errors, which is quite necessary for practical use.

\section{Experiment}

\subsection{Generative Relation Extraction}
Generative relation extraction tackles the conventional relation extraction problem by utilizing the seq2seq framework. At each time step, the model decoder either predicts the relation or copy a token from the input sentence. We focus on the task of extracting multiple tuples from sentences. We choose the New York Times (NYT) corpus\citep{zeng-etal-2018-extracting} for our experiments.  The detailed statistics are listed in Table~\ref{tab:accents}. Intuitively, the model capability of extracting long-span entities can be boosted by adding the proposed BioCopy. 

\begin{table}[h]
\small
\centering
\begin{tabular}{lcc}
\hline

& \textbf{Train} & \textbf{Test} \\\hline

examples  & 56,000 & 5,000 \\ 
triplets  & 88,366 & 8,120 \\ \hline
2 token &   37,352     & 3,335    \\ 
3 token &     6,362     &    566   \\
3+ tokens   &  1,259     &  112 \\ \hline

\end{tabular}
\small
\caption{Statistics of train/test split of the NYT. $n$-token denotes the number of examples that contain $n$-token entities.}
\label{tab:accents}
\end{table}

\paragraph{Baselines}
We compare our model performance with the following state-of-the-art relation extraction models.
\textbf{Tagging}\citep{zheng2017joint} is a neural sequence labeling model which jointly extracts the entities and relations using an LSTM encoder and an LSTM decoder. \textbf{CopyR}\citep{zeng-etal-2018-extracting} uses an encoder-decoder framework to extract entities and relations jointly. \textbf{GraphR}\citep{fu2019graphrel} models each token in a sentence as a node, and edges connecting the nodes as relations between them, they adopt the graph neural network to predict the relation triplets. \textbf{Ngram-att} uses an encoder-decoder framework with a n-gram attention layer. The encoder takes source sequence as input and decoder produces entity and relation IDs from Wikidata\citep{vrandevcic2014wikidata}. \textbf{WordDec} and \textbf{PointerDec} are both originated from~\citep{nayak2020effective}, where they use both word decoder and pointerNet as model decoder.

\paragraph{Model Details}
Since WordDec~\citep{nayak2020effective} is the state-of-the-art model on the NYT dataset, thus we select WordDec as our backbone model and all the model details are aligned with the original settings. Concretely, we initialize the word vectors by using Word2Vec\citep{mikolovdistributed}. We set the word embedding dimension $d_w$ = 300 and relation embedding dimension $d_r$ = 300, the hidden dimension $d_h$ of the LSTM cell is set at 300. The model is trained with the mini-batch size of 32 and the network parameters are optimized using Adam\citep{kingma2015adam}. Dropout layers with a dropout rate ﬁxed at 0.3 are used in our network to avoid overﬁtting.

\paragraph{Result}

\begin{table}[h]
\centering
\scalebox{0.81}{
\begin{tabular}{lccc}
\hline  \textbf{Models} & \textbf{Precision} & \textbf{Recall} & \textbf{F1 score}  \\ \hline
Tagging & 62.4\% & 31.7\% & 42.0\% \\
CopyR & 61.0\% & 56.6\% & 58.7\% \\
GraphR & 63.9\% & 60.0\% & 61.9\% \\
Ngram-att & 78.3\% & 69.8\% & 73.8\% \\
PointerDec & 80.6\% & 77.3\% & 78.9\% \\
WordDec & \textbf{88.1}\% & 76.1\% & 81.7\% \\
Our model & 87.7\% & \textbf{77.7}\% & \textbf{82.4}\% \\
\hline
\end{tabular}}
\small
\caption{\label{font-table} Performance comparison on NYT24 dataset}
\end{table}

We run the model with the above experimental settings, and we get the result as shown in Table \ref{font-table}. Our model outperforms the state-of-the-art methods 0.4\% by recall and 0.7\% by F1 score respectively. 

To further verify the effectiveness of the proposed BioCopy, We conduct the ablation experiment, and the results are presented in Table~\ref{t4}. To be more specific, we firstly use a raw seq2seq model without any involvement of copy mechanism. Then, we add the \textit{Attention Copy} that has been conducted by WordDec~\citep{nayak2020effective}, i.e., if the predicted token is unknown, we will select the token with the highest attention score from the input sentence . As we can see from Table~\ref{t4}, \textit{raw seq2seq} unsurprisingly turns out to be the lowest performance due to the massive prediction of unknown. Adding \textit{Attention Copy} can alleviate the unknown problem, but it is still a lack of capability of extracting long-span entities. By adding our \textit{BioCopy}, the model performance exceeds the other two since our model is able to not only deal with the unknown problem, but extracting the long-span entities properly.

\begin{table}[h]
	\centering
	\scalebox{0.81}{
	\begin{tabular}{lccc}
		\hline  & \textbf{Precision} & \textbf{Recall} & \textbf{F1 score}  \\ \hline
		Raw Seq2seq & 71.4\% & 59.6\% & 65.0\% \\
		+ Attention Copy & \textbf{88.1}\% & 76.1\% & 81.7\% \\
		+ BioCopy & 87.7\% & \textbf{77.7}\% & \textbf{82.4}\% \\
		\hline
	\end{tabular}}
	\caption{\label{t4} Model performance with different settings}
\end{table}
Table~\ref{t4} suggests our approach boosts the model performance to some extent. However, it is still not clear how effectively could our method act on multi-token entity extraction. Since the multi-token entities can be regarded as a long span, we speculate that our method can improve the model capability in terms of tackling the multi-token entities, We conduct an auxiliary experiment to verify this. As shown in Table~\ref{t5}, we can notice that the percentage of error cases, which are caused by incorrectly extracting long-span entities, drop from 49.1\% to 18.6\% by adding our BioCopy where it strongly verifies that the improvement of the model performance in this task is due to the reduction of long-span copy errors.

\begin{table}[h]
	\centering
	\scalebox{0.9}{
	\begin{tabular}{lc}
		\hline  & \textbf{Error Percentage}  \\ \hline

		Raw Seq2seq & 49.1\% \\
		+ Attention Copy & 23.7\% \\
		+ BioCopy & 18.6\% \\
		\hline
	\end{tabular}}
	\caption{\label{t5} Percentage of error cases of extracting long-span entities.}
\end{table}

\subsection{Auxiliary Experiment}

To verify the robustness of our proposed method, we also conduct an auxiliary experiment on abstractive summarization task. Abstractive summarization task aims to generate a new shorter text that conveys the most critical information from the original long text, where it usually requires to generate much longer text spans from the source sentenc, which can be leveraged to further evaluate our BioCopy.

We conducted our experiment on a Chinese legal summarization dataset (CAIL2020)~\footnote{https://github.com/china-ai-law-challenge/CAIL2020/tree/master/sfzy}, which contains a large number of legal terms. CAIL2020 dataset has 9,484 sample pairs. Each source text contains an average of 2569 words and each summary text contains 283 words.

\paragraph{Baseline and Metrics}
Recently, pre-training models have achieved promising results when fine-tuned on several text summarization tasks~\citep{dongunified, lewis2020bart}. We choose NEZHA~\citep{dongunified} as our backbone model, which is a large-scale Chinese pre-trained model and is able to encode the input text with any length. For a fair comparison, we convert the original encoder mask of NEZHA to the seq2seq mask as same as the BERT-UniLM~\citep{dong2019unified} used in Table~\ref{cail2020_results}. We use ROUGE scores for evaluation, in which Total is calculated as a weighted average of the above scores.

\paragraph{Result} 
In Table~\ref{cail2020_results}, we first compare the model performance with BERT-based models. We can notice that BERT model can achieve a better result by utilizing our BioCopy than by just simply using UniLM~\citep{dong2019unified}. Since the dataset is in Chinese, we select NEZHA~\citep{dongunified}
as our backbone model as it is pre-trained on massive Chinese corpus. The results in  Table~\ref{cail2020_results} show that the model can gain 71.31 Rouge-1, 54.72 Rouge-2, 70.29 Rouge-L and 64.29 Total, respectively, where it surpasses all the baseline models.
\begin{table}
	\centering
	\scalebox{0.81}{
	\begin{tabular}{lcccc}
		\hline  \textbf{Models} & \textbf{Rouge-1} & \textbf{Rouge-2} & \textbf{Rouge-L} & \textbf{Total} \\ \hline
		LSTM-Seq2seq & 46.48 & 30.48 & 41.80 & 38.21 \\
		BERT-UniLM & 63.83 & 51.29	& 59.76 & 57.19 \\
		BERT-BioCopy & 64.98 & 53.92 & 66.54 & 61.18 \\
		NEZHA & 70.93 & 54.38 & 69.89 & 63.89 \\
		NEZHA-BioCopy & \textbf{71.31} & \textbf{54.72} & \textbf{70.29} & \textbf{64.29}\\
		\hline
	\end{tabular}}
	\caption{\label{cail2020_results} Performance comparison on CAIL2020 dataset}
\end{table}

\section{Related Work}

\paragraph{Generative relation extraction.}\citep{zeng-etal-2018-extracting} proposed CopyRE, a joint model based on a copy mechanism, which transforms the joint extraction task into a generation task.\citep{nayak2020effective} propose two different encoders. The word decoder generates multiple triplets in a token-by-token manner, and each triplets is distinguished by the special token, while the pointer decoder simply utilizes pointerNet~\citep{2015arXiv150603134V} to generate start and end indexes for each entity. CopyMTL~\citep{2019arXiv191110438Z} introduced a multi-task framework with a sequence labeling layer in the encoder, to alleviate the problem that CopyRE can only extract the last token in a multi-token entity. 
\paragraph{Abstractive summarization.} The previous work mainly leverage an encoder-decoder framework
by choosing different model structures.~\citep{zhong2019closer} utilize Transformers or graph neural network~\citep{wang2020heterogeneous} for model encoder. Despite most of the previous work conducting the seq2seq model, some works~\citep{wang2019exploring}  deploy a non-autoregressive model decoder to tackle this task, which also shows great effectiveness.  

\section{Conclusion}

In this paper, we propose BioCopy, a plug-and-play copy mechanism to alleviate the long-span copying problem in generative tasks. By adding an extra sequence prediction layer in the training stage, our proposed approach is able to diminish the scope of probability distribution on each token. Experiments in generative relation extraction and abstractive summarization verifies the model's effectiveness. 
\bibliography{anthology,custom}
\bibliographystyle{acl_natbib}

\end{document}